%% file: main.tex
\documentclass[sigconf]{acmart}
\AtBeginDocument{%
  }

\setcopyright{acmlicensed}
\copyrightyear{2018}
\acmYear{2018}
\acmDOI{XXXXXXX.XXXXXXX}
\acmConference[KDD '26]{Proceedings of the 32nd ACM SIGKDD Conference on Knowledge Discovery and Data Mining}{August 9--13, 2026}{Jeju, Korea}

\acmISBN{978-1-4503-XXXX-X/2018/06}

\renewcommand\footnotetextcopyrightpermission[1]{}

\usepackage{array}
\usepackage{multirow}
\usepackage{tabularx}
\usepackage{xcolor}
\usepackage[most]{tcolorbox}
\usepackage{caption}
\usepackage{enumitem}

\setlist[itemize]{leftmargin=1.5em, topsep=0pt, partopsep=0pt}

\makeatletter
\providecommand{\captionof}[2]{\begingroup\def\@captype{#1}\caption{#2}\endgroup}
\makeatother

\definecolor{caseQHeader}{HTML}{7A7A7A}    
\definecolor{caseQFrame}{HTML}{555555}
\definecolor{caseOursHeader}{HTML}{AED6FC} 
\definecolor{caseOursFrame}{HTML}{0983F6}
\definecolor{caseOursTitle}{HTML}{000000}
\definecolor{caseBaseHeader}{HTML}{7EDDD2} 
\definecolor{caseBaseFrame}{HTML}{2A9D8F}
\definecolor{caseBaseTitle}{HTML}{000000}
\definecolor{caseAnaHeader}{HTML}{D7BDE2}  
\definecolor{caseAnaFrame}{HTML}{8E44AD}   
\definecolor{caseAnaTitle}{HTML}{000000}
\definecolor{textRed}{HTML}{D9534F}     
\definecolor{textGreen}{HTML}{2E7D32}   

\tcbset{
  casestudy/base/.style={
    enhanced,
    boxrule=0.5mm,
    arc=2mm,
    left=1.5mm,right=1.5mm,top=1.5mm,bottom=1.5mm,
    colback=white,
    fonttitle=\bfseries\small,
    fontupper=\small,
    before skip=1mm,
    after skip=1mm,
    before upper={\setlength{\parskip}{0.3em}}, 
  },
  casestudy/query/.style={
    casestudy/base,
    colframe=caseQFrame,
    colbacktitle=caseQHeader,
    coltitle=white,
    halign title=left,
  },
  casestudy/ours/.style={
    casestudy/base,
    colframe=caseOursFrame,
    colbacktitle=caseOursHeader,
    coltitle=caseOursTitle,
  },
  casestudy/baseline/.style={
    casestudy/base,
    colframe=caseBaseFrame,
    colbacktitle=caseBaseHeader,
    coltitle=caseBaseTitle,
  },
  casestudy/analysis/.style={
    casestudy/base,
    colframe=caseAnaFrame,
    colbacktitle=caseAnaHeader,
    coltitle=caseAnaTitle,
    title={Analysis \& Critique},
    fontupper=\small\itshape,
  }
}

\begin{document}

\title{Aligning Large Language Models with Searcher Preferences}

\author{Wei Wu}
\authornote{Equal Contribution}
\orcid{0009-0009-1590-601X}
\affiliation{%
  \institution{School of Artificial Intelligence and Data Science, University of Science and Technology of China}
  \city{Hefei}
  \country{China}
}
\email{urara@mail.ustc.edu.cn}

\author{Peilun Zhou}
\authornotemark[1]
\affiliation{%
  \institution{Xiaohongshu Inc.}
  \city{Beijing}
  \country{China}
}
\email{zhoupeilun@xiaohongshu.com}

\author{Liyi Chen}
\authornote{Corresponding Authors}
\affiliation{%
  \institution{Xiaohongshu Inc.}
  \city{Beijing}
  \country{China}
}
\email{chenliyi@xiaohongshu.com}

\author{Qimeng Wang}
\affiliation{%
  \institution{Xiaohongshu Inc.}
  \city{Beijing}
  \country{China}
}
\email{qimengwang@xiaohongshu.com}

\author{Chengqiang Lu}
\affiliation{%
  \institution{Xiaohongshu Inc.}
  \city{Beijing}
  \country{China}
}
\email{lunar@mail.ustc.edu.cn}

\author{Yan Gao}
\affiliation{%
  \institution{Xiaohongshu Inc.}
  \city{Beijing}
  \country{China}
}
\email{wanjianyi@xiaohongshu.com}

\author{Yi Wu}
\affiliation{%
  \institution{Xiaohongshu Inc.}
  \city{Beijing}
  \country{China}
}
\email{luyun2@xiaohongshu.com}

\author{Yao Hu}
\affiliation{%
  \institution{Xiaohongshu Inc.}
  \city{Beijing}
  \country{China}
}
\email{xiahou@xiaohongshu.com}

\author{Hui Xiong}
\orcid{0000-0001-6016-6465}
\authornotemark[2]
\affiliation{%
  \institution{Thrust of Artificial Intelligence, The Hong Kong University of Science and Technology (Guangzhou)}
  \city{Guangzhou}
  \country{China}
}
\affiliation{%
  \institution{Department of Computer Science and Engineering, The Hong Kong University of Science and Technology}
  \city{Hong Kong SAR}
  \country{China}
}
\email{xionghui@ust.hk}

\renewcommand{\shortauthors}{Wei Wu, Peilun Zhou, et al.}

\input{sections/0_abstract}


\maketitle
\input{sections/1_introduction}
\input{sections/2_relatedwork}
\input{sections/3_methodology}
\input{sections/4_experiments}
\input{sections/5_conclusion}

\bibliographystyle{ACM-Reference-Format}
\bibliography{base}

\appendix
\input{sections/6_appendix}

\end{document}

%% file: sections/0_abstract.tex
\begin{abstract}
The paradigm shift from item-centric ranking to answer-centric synthesis is redefining the role of search engines. While recent industrial progress has applied generative techniques to closed-set item ranking in e-commerce, research and deployment of open-ended generative search on large content platforms remain limited. This setting introduces challenges, including robustness to noisy retrieval, non-negotiable safety guarantees, and alignment with diverse user needs. In this work, we introduce SearchLLM, the first large language model (LLM) for open-ended generative search. We design a hierarchical, multi-dimensional reward system that separates bottom-line constraints, including factual grounding, basic answer quality and format compliance, from behavior optimization objectives that promote robustness to noisy retrieval and alignment with user needs. Concretely, our reward model evaluates responses conditioned on the user query, session history, and retrieved evidence set, combining rule-based checks with human-calibrated LLM judges to produce an interpretable score vector over these dimensions. We introduce a Gated Aggregation Strategy to derive the training reward for optimizing SearchLLM with Group Relative Policy Optimization (GRPO). We deploy SearchLLM in the AI search entry of RedNote. Offline evaluations and online A/B tests show improved generation quality and user engagement, increasing Valid Consumption Rate by 1.03\% and reducing Re-search Rate by 2.81\%, while upholding strict safety and reliability standards.
\vspace{-5pt}
\end{abstract}

%% file: sections/1_introduction.tex
\section{Introduction}
\label{sec:introduction}
The rapid growth in both the scale and diversity of digital content on online platforms has resulted in severe information overload~\citep{information-overload}, making it increasingly challenging for users to efficiently locate accurate answers and make informed decisions. Despite significant advances in search engines that have improved retrieval precision by providing more relevant items~\citep{llm-IR}, outputs of search engines typically remain as itemized lists. This format places a cognitive burden on users, who still need to further sift through and synthesize the results to fulfill their information needs. Recently, large language models (LLMs) have enabled a new open-ended paradigm of generative search systems~\citep{douyin,tiktok,wechat,gemini,chatgpt}, which can comprehensively assess the relevance and validity of retrieved information and directly generate coherent, natural-language answers to user queries. This transition from closed-set item retrieval to open-ended answer generation fundamentally redefines the role of search engines from merely returning candidate items to delivering synthesized, user-centered solutions.

Nowadays, open-ended generative search has been increasingly integrated into widely adopted real-world online platforms and is used by hundreds of millions of users, positioning it as a critical interface for large-scale information access and sensemaking.
Across short-video apps (e.g., Douyin~\citep{douyin}/TikTok~\citep{tiktok}), super-app ecosystems (e.g., WeChat~\citep{wechat}), and general-purpose assistants (e.g., Gemini~\citep{gemini}/ChatGPT~\citep{chatgpt}), these systems ground responses in heterogeneous sources---videos, live streams, hybrid posts, official accounts/news, and the web---and synthesize them into direct conversational answers, enabling faster access, clearer summaries, and more interactive exploration.
In the literature, industrial generative search systems are still largely item-centric and mainly focus on closed-set item generation in e-commerce domain, where the goal is to generate or rank product identifiers rather than produce free-form answers. For example, OneSearch~\citep{onesearch} enables production-scale generative retrieval via semantic ID tokenization with behavior-aware training; GRAM~\citep{gram} generates product IDs auto-regressively in a unified query--item code space; and CRS~\citep{crs} post-trains LLMs with SFT and RL to better capture user preferences for ranking.

\input{figures/tex/fig1}
\input{figures/tex/fig2}

However, such closed-set formulations are not directly applicable to open-ended generative search, where the system needs to synthesize grounded natural-language answers from heterogeneous and potentially noisy evidence under real-world user intents. As illustrated in Figure~\ref{fig:fig1}, we show the interface of open-ended generative search in the RedNote. The bottom-right subfigure further summarizes the attribution analysis of failure cases collected from online user feedback. Mitigating these issues requires aligning the underlying LLMs with three key requirements.
\textbf{R1: Robustness to noisy queries and evidence.} LLMs should remain robust under ambiguous or underspecified queries and noisy, heterogeneous, outdated, redundant, or even conflicting retrieved evidence, while deciding when to infer intent, ask for clarification, or refuse unsafe requests.
\textbf{R2: Bottom-line guarantees on reliability and safety.} Generated answers should satisfy strict reliability and safety guards, including factual grounding in provided evidence or widely accepted background knowledge, safety and policy compliance, logical consistency, and a controllable response format.
\textbf{R3: Alignment with user needs.} 
Answers should be optimized for user consumption and decision support, presenting key information early, reducing redundancy, and choosing an appropriate level of detail and structure for each query without weakening the constraints above.
Therefore, how to effectively train an LLM for open-ended generative search is of central importance.

In this work, we introduce SearchLLM, the first LLM for open-ended generative search. 
We first propose a reward system that encodes the three requirements above, and train SearchLLM under this objective.
Concretely, we consolidate the three requirements into a two-layer design that separates non-negotiable safeguards from user-facing quality objectives: a bottom-line layer that encodes R2 as hard constraints on factual grounding, safety, and response format, and a behavior optimization layer that, within this safe region, jointly captures R1 and R3 by shaping how the model responds to uncertain queries and noisy evidence and how it trades off brevity, coverage, and novelty. On top of this structural design, we instantiate a hybrid evaluation stack that combines deterministic rule-based evaluators with LLM-based judges, each designed to output fine-grained, interpretable scores on specific sub-dimensions of the metric space. We calibrate our evaluators via a human-in-the-loop process so that the resulting interpretable metrics remain faithful to searcher preferences and stable in deployment. The resulting multi-dimensional reward vector is then transformed by a Gated Aggregation Strategy that protects bottom-line dimensions and rebalances user-facing quality dimensions, and is subsequently used as the objective in a Group Relative Policy Optimization~\citep{grpo} (GRPO)-style reinforcement learning pipeline trained over large-scale search query logs. We deploy SearchLLM in the AI search entry of RedNote. Online A/B tests demonstrate that our method yields measurable gains in user satisfaction, specifically increasing the Valid Consumption Rate (VCR) by 1.03\% and reducing the Re-search Rate (RR) by 2.81\% compared to the production baseline.
We summarize our main contributions as follows.
\begin{itemize}
  \item We are the first to characterize the unique demands of open-ended generative search on large content platforms and introduce a dedicated LLM named SearchLLM, which can provide valuable insights for the community.
  \item We propose a multi-dimensional reward design for generative search that explicitly separates non-negotiable safeguards from user-facing quality objectives, and instantiate it with a hybrid stack of rule-based checks and LLM judges calibrated by experts.
  \item We develop an end-to-end reinforcement learning recipe for SearchLLM, optimizing the full workflow with GRPO-style training and a Gated Aggregation Strategy that prioritizes bottom-line constraints while improving robustness and utility.
  \item We deploy SearchLLM in RedNote’s AI search entry, and online A/B tests show improved user experience, with +1.03\% Valid Consumption Rate (VCR) and -2.81\% Re-search Rate (RR) over the production baseline.
\end{itemize}

%% file: figures/tex/fig1.tex
\begin{figure}[t!]
  \centering
  \includegraphics[width=\linewidth]{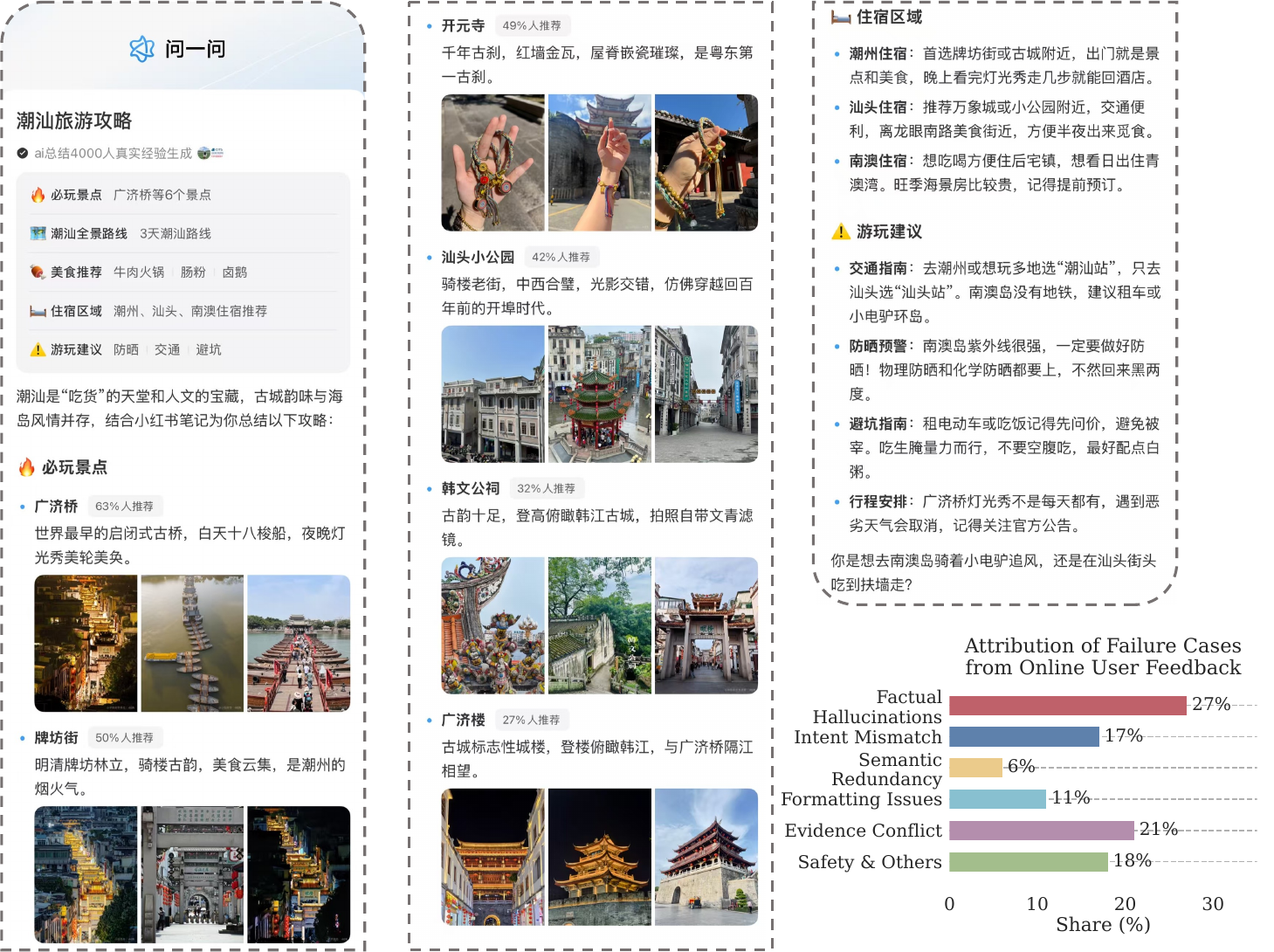}
  \vspace{-20pt}
  \caption{User interaction snapshots of open-ended generative search in RedNote. The bottom-right panel summarizes failure attribution from online user feedback.}
  \label{fig:fig1}
  \vspace{-14pt}
\end{figure}

%% file: figures/tex/fig2.tex
\begin{figure*}[t!]
  \centering
  \includegraphics[width=\linewidth]{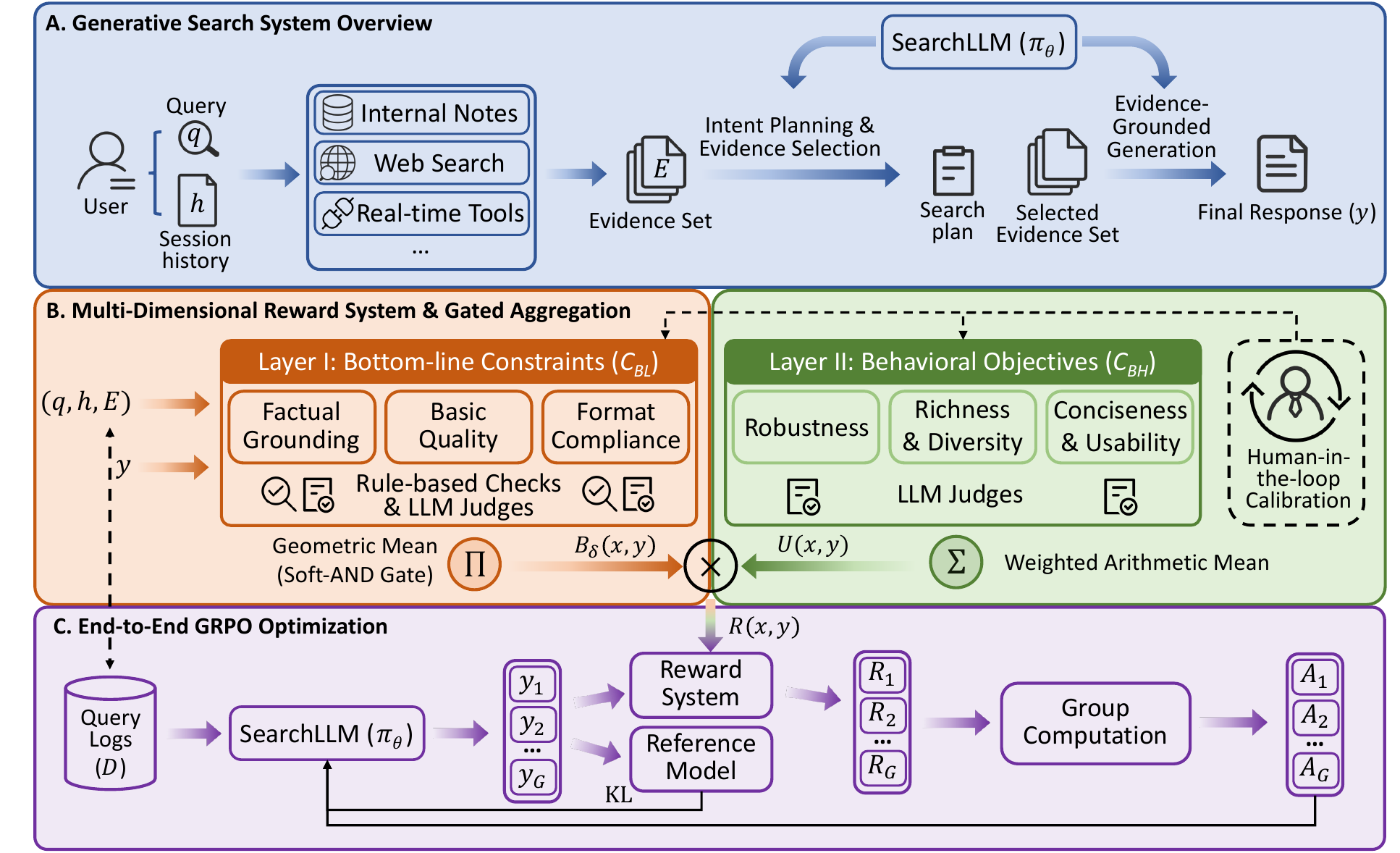}
  \vspace{-18pt}
  \caption{Overview of the alignment framework for open-ended generative search. The pipeline incorporates a multi-dimensional reward system that explicitly decouples non-negotiable bottom-line constraints (Layer I) from behavioral optimization objectives (Layer II). A hybrid evaluation stack, consisting of deterministic rules and human-calibrated LLM judges, computes fine-grained scores across multiple dimensions. These signals are synthesized via a gated aggregation mechanism to stabilize the learning signal for Group Relative Policy Optimization (GRPO).}
  \label{fig:fig2}
  \vspace{-8pt}
\end{figure*}

%% file: sections/2_relatedwork.tex
\vspace{-2pt}
\section{Related Work}
\paragraph{\textbf{Large Language Models for Search}}
LLMs are gradually reshaping how search engines operate and interact with users~\citep{search-meet-llm}.
At the component level, LLMs are extensively used to enhance query understanding via rewriting and expansion~\citep{query2doc,query-rewrite-1}, improve document indexing through semantic embedding~\citep{m3-embedding}, and refine ranking by estimating relevance more accurately~\citep{llm-ranker,re-ranking}. 
Beyond these component-level optimizations, recent industrial efforts in the e-commerce domain have explored end-to-end generative retrieval. 
Notable systems like OneSearch~\citep{onesearch} employ semantic ID tokenization to perform generative retrieval at scale, while GRAM~\citep{gram} and CRS~\citep{crs} generate product identifiers auto-regressively in a unified query--item latent space. 
In parallel, the paradigm of retrieval-augmented generation (RAG)~\citep{rag-survey,llm-IR} has shifted the focus from returning ranked lists to synthesizing natural-language answers grounded in retrieved evidence~\citep{generative-search-1,generative-search-2}.
However, in the literature, industrial applications predominantly focus on closed-set item ranking for e-commerce. Systematic studies on deploying open-ended generative search remain limited at scale.

\vspace{-2pt}
\paragraph{\textbf{Alignment of Large Language Models}}
Aligning LLMs with human values and safety standards is critical for deployment. 
Standard approaches typically utilize Reinforcement Learning from Human Feedback (RLHF) via Proximal Policy Optimization (PPO)~\citep{ppo,instructgpt}, or employ more efficient offline variants such as Direct Preference Optimization (DPO)~\citep{dpo} and Group Relative Policy Optimization (GRPO)~\citep{grpo}.
Regardless of the optimization algorithm, the efficacy of alignment relies on the design of the reward signal, which falls into three categories: 
(i) Scalar reward models~\citep{instructgpt,internlm2,GRM}, which map responses to a single quality score. While computationally efficient, they produce coarse-grained, opaque scores that act as black boxes, making it difficult to balance competing objectives like safety and helpfulness;
(ii) LLM-as-a-judge approaches~\citep{llm-as-a-judge,rlaif}, which leverage powerful LLMs to scale up evaluation without human labelers, though they can be sensitive to prompting and inherent model biases; and 
(iii) Rubric-based rewards~\citep{deliberative-alignment,omni-thinker,checklists-are-better}, which decompose complex human preferences into structured, interpretable criteria (e.g., checklists). 
Building on these advances, we propose a two-layer, multi-dimensional reward system for open-ended generative search that explicitly separates non-negotiable safeguards from user-facing quality objectives, instantiated via a hybrid stack of rule-based checks and calibrated LLM judges.

%% file: sections/3_methodology.tex
\vspace{-4pt}
\section{Methodology}
\subsection{System Overview}
We design an LLM-based open-ended generative search pipeline (illustrated in Figure~\ref{fig:fig2}) that enhances retrieval-augmented generation by using a single LLM for \textit{intent planning}, \textit{evidence selection}, and \textit{evidence-grounded generation} in a unified workflow. Built on top of a standard multi-source retrieval infrastructure that provides a broad pool of candidate evidence from internal note search, web search, and real-time tools, our system focuses on reasoning over this candidate pool rather than on retrieval itself. Given a user query and its session history, it first constructs an explicit intent plan that structures the information need into several interpretable facets and specifies how each facet should be supported by evidence. Conditioned on this plan, the system then performs intent-aware evidence selection that filters the candidate pool into a compact evidence set that is relevant, non-redundant, and collectively covers the planned facets. This curated evidence set is finally consumed by the LLM to synthesize a grounded final response.

\vspace{-4pt}
\subsection{Multi-Dimensional Reward System}
\label{sec:3_2}
Our reward system does not evaluate responses in isolation, but instead scores complete search interactions. Formally, let $x = (q, h, E)$ denote the input context comprising the query $q$, session history $h$, and retrieved evidence set $E$, and let $y$ denote the generated sequence, which encompasses both the intent plan and the final answer. 
To operationalize the critical requirements identified in Section \ref{sec:introduction}, specifically the inherent conflict between enforcing strict bottom-line guarantees (R2) and optimizing for robustness and user alignment (R1, R3), we move beyond monolithic scalar reward models. Drawing inspiration from recent work on rubric-based rewards~\citep{deliberative-alignment,rubrics-as-rewards,checklists-are-better,omni-thinker}, we design a hierarchical and multi-dimensional reward system. This system explicitly decouples \textit{bottom-line constraints}, which ensure non-negotiable safety and reliability, from \textit{behavioral optimization objectives}, which target robustness and preference alignment. This separation ensures that the model prioritizes foundational guarantees while simultaneously optimizing the user's search experience across the entire pipeline, from intent planning and evidence selection to final response synthesis.

\vspace{-4pt}
\subsubsection{Reward Design}
We formalize the evaluation of a model generation $y$ for a given search interaction context $x$ through a structured set of evaluation criteria $\mathcal{C}$. Unlike standard RLHF~\cite{ppo} which often conflates diverse quality signals into a single scalar, we categorize our criteria into two distinct layers:
\vspace{-4pt}
\paragraph{Layer I: Bottom-line Constraints.}
This layer encodes Requirement R2 (Reliability and Safety). It consists of binary or near-binary criteria that serve as hard constraints. Failure in these dimensions renders a response unusable. We define a subset of criteria $\mathcal{C}_{BL} \subset \mathcal{C}$ covering three critical aspects:
\begin{itemize} 
\item \textbf{Hallucination \& Factual Grounding:} To mitigate misinformation, we rigorously assess hallucination at both the sentence and claim levels, ensuring that the generated answer is factually accurate. Additionally, we verify consistency with external knowledge bases and enforce strict refusal behaviors when retrieved evidence is insufficient to answer the query.
\item \textbf{Basic Answer Quality:} This subset ensures the fundamental logic and readability of the response. It detects logical inconsistencies within a single turn, contradictions across multi-turn interactions, and filters out gibberish or low-quality text.
\item \textbf{Format Compliance:} To ensure the response is structurally consumable, we enforce strict formatting rules, including adherence to Markdown and constraints on response length.
\end{itemize}

\vspace{-4pt}
\paragraph{Layer II: Behavioral Objectives.}
Once bottom-line constraints are satisfied, the model should be optimized for robustness (R1) and user utility (R3). This layer, denoted as $\mathcal{C}_{BH} \subset \mathcal{C}$, specifically targets increasing user engagement with the generated responses. Key dimensions include:

\begin{itemize} 
\item \textbf{Robustness to Query \& Evidence:} This subset focuses on the complexity of open-ended generative search. On the query side, we evaluate intent alignment to ensure the model addresses the user's core question. On the evidence side, we assess how useful the utilized content is for answering the query and how likely it is to satisfy the user, while penalizing the inclusion of irrelevant or conflicting information. Furthermore, we evaluate the quality of planning and reasoning under uncertainty.
\item \textbf{Richness \& Diversity:} To enhance the depth of information and avoid overly narrow answers, we encourage the generation of diverse yet relevant claims, ensuring the response covers multiple perspectives or facets of the topic rather than repeating a single point.
\item \textbf{Conciseness \& Usability:} To optimize the consumption experience, we prioritize usability metrics. This includes the "answer-first" principle (placing the core answer at the beginning), reducing semantic redundancy, and minimizing off-topic or tangential content to ensure a high signal-to-noise ratio.
\end{itemize}
Table~\ref{tab:table5} (Appendix~\ref{app:reward}) provides the detailed definitions for all criteria discussed above, alongside their specific implementation types which we detail in the following section.

\vspace{-4pt}
\subsubsection{Implementation of Hybrid Evaluation Stack}
To ensure our reward signals are both scalable and diagnostically precise, we operationalize a hybrid evaluation stack that integrates deterministic rules (e.g., n-gram statistics, regex constraints) for objective criteria with LLM-based judges for complex semantic dimensions. Aligning these LLM judges with human expert preferences is non-trivial; strictly relying on single-pass annotation often leads to noisy ground truth. To address this, we establish a rigorous human-in-the-loop calibration cycle. Figure~\ref{fig:fig3} shows the annotation interface used for this process, where experts provide both fine-grained scores and holistic pairwise rankings. To strictly control annotation quality and mitigate cognitive inertia (anchoring bias) where annotators might over-trust the model's generated reasoning, we enforce a dual-track annotation protocol involving two distinct groups:
\begin{itemize}
    \item \textbf{Blind Group:} Annotators score responses based on the query and answer text, without access to the reward system's internal reasoning or intermediate steps. This establishes an unbiased baseline that reflects the end-user's consumption experience.
    \item \textbf{Assisted Group:} Annotators review the response alongside the reward system's CoT and evidence citations. This setup enables them to verify logical consistency and detect subtle hallucinations that are factually incorrect but linguistically plausible.
\end{itemize}
Discrepancies between the Blind and Assisted scores serve as high-value signals for identifying policy gaps; these conflict cases are escalated to senior experts for adjudication to resolve edge-case ambiguities and unify evaluation standards. Finally, before any version update, the evaluator must demonstrate non-degradation on a frozen regression benchmark and undergo a stable "shadow run" in the production environment.

\vspace{-4pt}
\subsection{Reinforcement Learning Framework}
Optimizing policy models against such a composite and fine-grained reward system presents a unique challenge. A naive linear combination of these diverse signals often exacerbates the "seesaw effect"~\citep{rubrics-as-rewards}, where the model exploits easier-to-optimize reward signals (e.g., length-based scores) at the expense of rigid bottom-line constraints such as factual consistency. To robustly operationalize this multi-dimensional feedback without succumbing to reward hacking, we introduce a \textit{Gated Aggregation Strategy}. 
For a given model generation $y$, let $S_{BL} = \{s_1, \ldots, s_m\}$ be the normalized scores from the bottom-line criteria defined in Layer~I, and let $S_{BH} = \{s'_1, \ldots, s'_n\}$ be the normalized scores from the behavioral criteria defined in Layer~II, with associated expert-defined weights $W = \{w_1, \ldots, w_n\}$. We assume that each score lies in the unit interval, \textit{i.e.}\ $s_i \in [0,1]$ for $i=1,\dots,m$ and $s'_i \in [0,1]$ for $i=1,\dots,n$.
We compute the bottom-line factor via a $\delta$-smoothed geometric mean
\begin{small}
\begin{equation}
  B_{\delta}(x,y)
  =
  \exp\left(
    \frac{1}{m}\sum_{i=1}^{m}
    \log\frac{s_i+\delta}{1+\delta}
  \right),
  \qquad \delta>0,
\label{eq:B}
\end{equation}
\end{small}which acts as a soft-AND gate while improving optimization stability. In the vanilla form (i.e., $\delta=0$),
\begin{small}
\[
  B_{0}(x,y)
  =
  \left(\prod_{i=1}^{m} s_i\right)^{1/m}
  =
  \exp\left(\frac{1}{m}\sum_{i=1}^{m}\log s_i\right), 
\]
\end{small}the gate can become numerically unstable when computed in the log-domain due to $\log 0$ and exhibits unbounded sensitivity
\begin{small}
\[
  \frac{\partial \log B_{0}(x,y)}{\partial s_i}=\frac{1}{m s_i},
\]
\end{small}when $s_i\to 0$. The $\delta$-smoothing avoids $\log 0$ and upper-bounds the gate sensitivity:
\begin{small}
\[
  \frac{\partial \log B_{\delta}(x,y)}{\partial s_i}
  =
  \frac{1}{m(s_i+\delta)}
  \le
  \frac{1}{m\delta}. 
\]
\end{small}Moreover, when any bottom-line metric is near zero, $B_{\delta}$ remains strongly suppressed in a $\delta$-controlled manner, yielding a substantially reduced reward. Conversely, we compute the behavioral utility as a weighted arithmetic mean
\begin{small}
\begin{equation}
  U(x,y) = \frac{\sum_{i=1}^{n} w_i s'_i}{\sum_{i=1}^{n} w_i},
  \label{eq:U}
\end{equation}
\end{small}which allows for flexible trade-offs among softer objectives within the safe region. 
The final scalar reward is then defined as
\begin{small}
\begin{equation}
  R(x,y) = B_{\delta}(x,y)\,U(x,y),
  \label{eq:reward}
\end{equation}
\end{small}so that behavioral improvements act as multipliers when bottom-line scores are sufficiently high.
To optimize the policy under this reward surface, we employ Group Relative Policy Optimization (GRPO)~\citep{grpo}. Unlike standard PPO~\citep{ppo} which relies on a value network, GRPO normalizes advantages within a group of sampled outputs for the same query, making it efficient for our large-scale setting. Formally, let $\pi_\theta$ denote the policy to be optimized and $\pi_{\theta_{\text{old}}}$ the behavior policy used to collect trajectories. For a query $x$, we sample a group of generations $\{y_i\}_{i=1}^{G}$, where each $y_i$ is a token sequence $y_i = (y_{i,1},\dots,y_{i,|y_i|})$. The GRPO objective is
\begin{small}
\begin{equation}
\begin{aligned}
    \mathcal{J}_{\text{GRPO}}(\theta)
    &= \mathbb{E}_{x \sim D,\ \{y_i\}_{i=1}^{G} \sim \pi_{\theta_{\text{old}}}(\cdot \mid x)}
    \Biggl[
      \frac{1}{G} \sum_{i=1}^{G} \frac{1}{|y_i|} \sum_{t=1}^{|y_i|}
      \biggl( \min\Bigl(
        r_{i,t}(\theta)\,\hat{A}_{i,t},\\
        &\qquad\quad\;\;
        \operatorname{clip}\bigl(r_{i,t}(\theta), 1-\epsilon, 1+\epsilon\bigr)\,\hat{A}_{i,t}
      \Bigr) - \lambda D_{\text{KL}}(\pi_\theta || \pi_{\text{ref}}) \biggr)
    \Biggr],
\end{aligned}
\label{eq:grpo}
\end{equation}
\end{small}where $\lambda$ is the coefficient controlling the KL divergence.
\begin{small}
\begin{equation}
  r_{i,t}(\theta)
  = 
  \frac{\pi_\theta\bigl(y_{i,t} \mid x, y_{i,<t}\bigr)}
       {\pi_{\theta_{\text{old}}}\bigl(y_{i,t} \mid x, y_{i,<t}\bigr)},
  \label{eq:is}
\end{equation}
\end{small}is the per-token importance-weighted ratio. The token-wise advantages $\hat{A}_{i,t}$ are obtained by normalizing the group rewards:
\begin{small}
\begin{equation}
  \hat{A}_{i,t}
  =
  \frac{R(x,y_i) -
        \mathrm{mean}\bigl(\{R(x,y_j)\}_{j=1}^{G}\bigr)}
       {\mathrm{std}\bigl(\{R(x,y_j)\}_{j=1}^{G}\bigr)},
  \;
  t = 1,\dots,|y_i|.
  \label{eq:advantage}
\end{equation}
\end{small}

%% file: sections/4_experiments.tex
\section{Experiments}
\input{tables/table12}
\input{tables/table3}
In this section, we evaluate the effectiveness of our proposed reward system and RL training framework. We aim to answer the following research questions:
\begin{itemize}
    \item \textbf{RQ1:} How effectively does our multi-dimensional reward system align with human expert judgments compared to state-of-the-art reward modeling baselines?
    \item \textbf{RQ2:} Does the proposed optimization strategy yield better offline generation quality?
    \item \textbf{RQ3:} How does the Gated Aggregation Strategy influence training dynamics and coordinate different reward dimensions?
    \item \textbf{RQ4:} What is the impact of the deployed model in online settings on real-world user engagement and safety metrics?
\end{itemize}

\vspace{-2pt}
\subsection{Experimental Setup}
\subsubsection{Datasets}
To support the training and evaluation of our framework, we construct four distinct datasets derived from RedNote search logs: a \textbf{Reward Training Dataset ($\mathcal{D}_{\text{RM-Train}}$)} for calibration, two diagnostic test sets (\textbf{$\mathcal{D}_{\text{Eval-Dim}}$} and \textbf{$\mathcal{D}_{\text{Eval-Holistic}}$}) for reward validation, and a large-scale unlabeled \textbf{RL Optimization Dataset ($\mathcal{D}_{\text{RL}}$)}. Detailed statistics and construction protocols for these datasets are provided in Appendix~\ref{app:datasets}.

\vspace{-2pt}
\subsubsection{Baselines}
To validate our contributions, we benchmark our approach against state-of-the-art methods from two perspectives: reward quality (RQ1) and policy optimization effectiveness (RQ2--4).

\vspace{-4pt}
\paragraph{Reward Modeling Baselines.} We compare our reward system against the most advanced reward modeling paradigms:
\begin{itemize}
    \item \textbf{GenRM (Generative Reward Model):} This baseline trains an LLM to generate a Chain-of-Thought (CoT) reasoning trace before outputting final scores~\cite{GRM,Generative-Verifiers}.
    \item \textbf{Rubric (Rubric-Based Reward):} Following~\citep{AdvancedIF}, this baseline trains a rubric generator via supervised fine-tuning on a small set of expert-written query--rubric pairs, then uses it to generate query-specific rubrics for each reward dimension to compute the rubric-based reward.
\end{itemize}

\vspace{-4pt}
\paragraph{Policy Baselines.} All policy models are initialized from the same SFT checkpoint, which was trained on high-quality business data to ensure basic service capability. We compare the following optimization strategies:
\begin{itemize}
    \item \textbf{RFT (Rejection Sampling Fine-tuning):} An iterative alignment method where the model generates multiple candidates per query, and the highest-scoring are selected to build a new dataset for this round of SFT~\citep{RFT1,RFT2}.
    \item \textbf{DPO (Direct Preference Optimization):} An offline method that optimizes the policy directly on preference data~\citep{dpo}, bypassing the explicit reward modeling step.
    \item \textbf{GRPO-GenRM:} GRPO~\citep{grpo} using the trained GenRM’s outputs as the reward signal.
    \item \textbf{GRPO-Linear:} A variant of our framework that utilizes our multi-dimensional reward signals but aggregates them via a naive weighted sum.
    \item \textbf{GRPO-Gated (Ours):} Our proposed method utilizing the Gated Aggregation Strategy.
\end{itemize}

\vspace{-2pt}
\subsubsection{Evaluation Metrics}
We employ a comprehensive set of metrics to assess both reward alignment and real-world performance. For offline evaluation, we utilize \textbf{Accuracy (ACC)} and \textbf{Area Under the Curve (AUC)} to measure agreement with human experts. For online deployment, we monitor user engagement via \textbf{Valid Consumption Rate (VCR)}, \textbf{Skip Rate (SR)}, \textbf{Re-search Rate (RR)}, and \textbf{Bad Case Rate (BCR)}. Detailed definitions and mathematical formulations for these metrics are provided in Appendix~\ref{app:metrics}.

\vspace{-4pt}
\subsubsection{Implementation Details}
All experiments are conducted on 18 nodes equipped with NVIDIA H800 GPUs where 16 nodes are dedicated to the reward system deployment. We implement the training pipeline using the verl~\citep{verl} library where we leverage Megatron~\citep{megatron} for distributed training and vLLM~\citep{vllm} for efficient generation rollouts. The policy model is initialized from Qwen3-30B-A3B-Instruct-2507~\cite{qwen3} while the entire reward evaluation stack is powered by DeepSeek-R1~\citep{r1} to ensure reasoning-dense feedback. We adopt an on-policy optimization configuration with a global batch size of 128 and sample 16 completions per prompt at a temperature of 1.0. The optimization utilizes the AdamW~\citep{adamw} optimizer with $\beta_1=0.9$ and $\beta_2=0.95$ plus a weight decay of 0.1 alongside a learning rate of $1 \times 10^{-6}$ and a KL-divergence coefficient $\lambda$ of 0.01. For the Gated Aggregation in Eq.~\ref{eq:B}, we set $\delta=10^{-2}$.

\vspace{-2pt}
\subsection{Validation of Reward System (RQ1)}
To evaluate the alignment of our reward system with human expert preferences, we compare on the $\mathcal{D}_{\text{Eval-Dim}}$ and the $\mathcal{D}_{\text{Eval-Holistic}}$.

\vspace{-2pt}
\subsubsection{Dimension-Specific Alignment}
\input{tables/table4}
Tables~\ref{tab:table1} and~\ref{tab:table2} show that our hybrid evaluation stack consistently outperforms standard GenRM and Rubric-based baselines across both bottom-line and behavioral layers. In safety and factual grounding (Table~\ref{tab:table1}), our system achieves superior accuracy. Although the Rubric-based method improves upon GenRM by decomposing complex tasks into explicit criteria to reduce holistic scoring ambiguity, it still falls short of our approach. This gap persists because standard Rubrics struggle with ambiguous instructions and content variability, whereas our human-in-the-loop calibration ensures strict alignment with expert consensus. This enables us to enforce bottom-line guarantees that pure LLM judges occasionally miss. Furthermore, GenRM exhibits lower stability due to its high sensitivity to training data distribution, which leads to inconsistent performance on out-of-distribution safety violations. Table~\ref{tab:table2} demonstrates our pairwise ranking capability for subjective behavioral objectives, where our approach shows a significant advantage in Robustness and Conciseness. Decoupling these sub-dimensions into interpretable criteria within a human-calibrated hybrid stack allows our system to provide more stable and fine-grained signals. In contrast, while GenRM generates reasoning traces, the Rubric baseline offers more structured diagnosis through explicit criteria yet lacks the continuous expert calibration loop required to align precisely with nuanced human preferences in open-ended generative search.

\vspace{-2pt}
\subsubsection{Holistic Preference Alignment}
\input{figures/tex/fig4}
Table~\ref{tab:table3} presents holistic alignment results on $\mathcal{D}_{\text{Eval-Holistic}}$. Our multi-dimensional reward system achieves the highest overall agreement with human preferences and surpasses GenRM and the Rubric-based baseline. This result validates the effectiveness of our two-layer design separates non-negotiable safeguards from user-facing quality objectives to capture the complexity of human judgments. The consistent superiority across all subsets confirms that the proposed reward system serves as a reliable objective for the GRPO-based policy optimization.

\vspace{-2pt}
\subsection{Offline Policy Evaluation (RQ2)}
To verify whether our proposed optimization strategy translates better reward signals into superior generation quality, we conduct a comprehensive offline evaluation. Relying on the high alignment of our reward system with human judgment demonstrated in RQ1, we employ the system itself as an automated proxy evaluator alongside rigorous human expert assessment.

\vspace{-2pt}
\subsubsection{Automatic Evaluation via Reward System}
\input{figures/tex/fig5}
\input{figures/tex/fig6}
Table~\ref{tab:table4} presents the performance of different policy variants evaluated by our multi-dimensional reward system. To facilitate comparison, we report average scores of dimensions within each subset. The results indicate that GRPO-Gated (Ours) achieves the best overall performance, significantly surpassing the SFT baseline and other optimization methods. While RFT and DPO improve upon the SFT baseline, they exhibit limitations in balancing conflicting objectives. Specifically, DPO tends to exploit easy-to-optimize patterns like response length while often failing to strictly adhere to complex safety constraints. Among the GRPO variants, GRPO-GenRM shows instability in safety dimensions due to the lack of hard constraints in its reward signal. GRPO-Linear, utilizing a weighted sum aggregation, suffers from the seesaw effect where improvements in behavioral scores come at the cost of degradation in bottom-line metrics. In contrast, our Gated Aggregation Strategy ensures that the model optimizes behavioral utility only within the safe region, resulting in simultaneous improvements across both safety and utility dimensions.

\vspace{-2pt}
\subsubsection{Human Expert Evaluation}
To corroborate the automatic metrics, we conducted a blind side-by-side human evaluation. We sampled 1,000 distinct queries from the test set and engaged a large pool of expert annotators to compare the responses generated by our method against representative baselines. The experts judged the responses based on critical criteria such as factual consistency, safety, richness, conciseness, and logic. The evaluation results (Figure~\ref{fig:fig4}) align consistently with the automatic metrics, showing that our model achieves a significantly higher win rate. Experts observed that our method produces answers that are not only robust to noisy retrieval and semantically rich but also strictly compliant with formatting and safety guidelines. This confirms that the gains observed in the reward scores genuinely reflect improvements in generation quality perceptible to human users.

\vspace{-2pt}
\subsection{Training Dynamics (RQ3)}
\label{sec:4_4}
We analyze the evolution of different reward dimensions during training, as illustrated in Figure~\ref{fig:fig5}, to understand how the Gated Aggregation Strategy resolves the inherent antagonism between conflicting objectives. In the early stages of training, we observe a natural seesaw effect in the baseline methods, particularly between Richness and Conciseness; as the model strives to cover more information, it inevitably tends towards verbosity, causing conciseness scores to drop. GRPO-Linear partially mitigates this issue by introducing expert priors through weighted summation, which forces the model to respect the relative importance of each dimension. However, it still treats safety constraints and behavioral objectives as parallel signals, leading to instability where the model occasionally sacrifices Bottom-line Constraints to chase marginal gains in utility. In contrast, GRPO-Gated fundamentally alters this dynamic by establishing a hierarchical dependency. The training curves show that the Gated strategy first "locks in" high scores on bottom-line constraints. Only after these safety metrics stabilize above the threshold does the model begin to effectively optimize for richness and diversity. This mechanism transforms the inter-dimensional conflict into a synergistic evolution, ensuring that the pursuit of user utility never compromises the foundational reliability of the generative search system.

\vspace{-2pt}
\subsection{Online Experiments (RQ4)}
To evaluate real-world impact and robustness, we deployed the trained model in the AI search entry of RedNote. We implemented randomized traffic diversion by hashing User IDs and modulo bucketing. We allocated a consistent 10\% of the total live traffic to each experimental group. All variants were evaluated synchronously over the same period to eliminate temporal fluctuations, thereby ensuring the fairness and reliability of the statistical conclusions. All reported uplifts are statistically significant (two-sided tests; $p<0.05$). In particular, for VCR the typical 95\% confidence interval is within $\pm 0.1$ percentage points, indicating that the observed gains are well above the noise level. Additionally, we explicitly excluded a small subset of vertical business domains from the training set to assess the generalization capability of the model across unseen distributions. The online A/B testing results, summarized in Figure~\ref{fig:fig6}, demonstrate substantial improvements across all core engagement metrics. Compared to the baseline, our model achieves a significant increase in Valid Consumption Rate (VCR) which indicates that users find the synthesized answers helpful and are willing to spend time reading them. Simultaneously, we observe a marked reduction in Skip Rate (SR) and Re-search Rate (RR). These shifts suggest that the generated responses satisfy user needs in a single turn and reduce the necessity for query reformulation. Crucially, the Bad Case Rate (BCR) remains at a minimal level which ensures the safety and reliability of the deployed system. Furthermore, the performance on the held-out domains confirms the robust generalization of our method. Although different business verticals exhibit distinct data biases regarding content format or information density, our model adapts effectively without explicit fine-tuning. It maintains high safety standards and response quality even in these zero-shot scenarios. This stability validates that our multi-dimensional reward system captures fundamental search principles rather than overfitting to specific training data patterns.

%% file: tables/table12.tex
\begin{table*}[t]
    \centering
    \caption{Comparison on pointwise constraint of our reward system against multiple baselines. Metric reported is Accuracy.}
    \vspace{-6pt}
    \label{tab:table1}
    \small
    \begin{tabular}{p{1.8cm} p{0.8cm} p{1.6cm} p{3.2cm} >{\centering\arraybackslash}p{1.2cm} >{\centering\arraybackslash}p{1.2cm} >{\centering\arraybackslash}p{1.2cm}}
        \toprule
        \textbf{Layer} & \textbf{Req.} & \textbf{Subset} & \textbf{Reward Dimension} & \textbf{GenRM} & \textbf{Rubric} & \textbf{Ours} \\
        \midrule
        \multirow{3}{*}{Behavioral} 
        & \multirow{3}{*}{R1} 
          & Query & Query Satisfy & 71.52 & 73.31 & \textbf{87.24} \\
          \cmidrule{3-7}
          & & \multirow{2}{*}{Evidence} & Reference Conflict & 79.01 & 70.48 & \textbf{90.84} \\
          & & & Reference Irrelevant & 88.92 & 83.63 & \textbf{94.96} \\
        \midrule
        \multirow{6}{*}{Bottom-line} 
        & \multirow{6}{*}{R2} 
          & \multirow{3}{*}{\shortstack[l]{Basic\\Quality}} & Answer Quality & 64.02 & 65.69 & \textbf{82.43} \\
          & & & Self Consistency & 59.68 & 64.22 & \textbf{78.20} \\
          & & & Multi-turn Repeat/Conflict & 80.60 & 83.40 & \textbf{94.20} \\
          \cmidrule{3-7}
          & & \multirow{3}{*}{\shortstack[l]{Halluci-\\nation}} & Highlight Hallucination & 49.34 & 69.05 & \textbf{92.85} \\
          & & & Claim Hallucination & 52.93 & 66.84 & \textbf{91.88} \\
          & & & LLM Knowledge & 84.19 & 84.19 & \textbf{85.77} \\
        \bottomrule
    \end{tabular}
    \vspace{-4pt}
\end{table*}

\begin{table*}[t]
    \centering
    \caption{Comparison on pairwise preference of our reward system against multiple baselines. Metric reported is AUC.}
    \vspace{-6pt}
    \label{tab:table2}
    \small
    \begin{tabular}{p{1.8cm} p{0.8cm} p{1.6cm} p{3.2cm} >{\centering\arraybackslash}p{1.2cm} >{\centering\arraybackslash}p{1.2cm} >{\centering\arraybackslash}p{1.2cm}}
        \toprule
        \textbf{Layer} & \textbf{Req.} & \textbf{Subset} & \textbf{Reward Dimension} & \textbf{GenRM} & \textbf{Rubric} & \textbf{Ours} \\
        \midrule
        \multirow{8}{*}{Behavioral} 
        & \multirow{4}{*}{R1} 
          & Query & Planning Quality & 84.74 & 73.68 & \textbf{89.47} \\
          \cmidrule{3-7}
          & & \multirow{3}{*}{Evidence} & Reference Beneficiality & 79.56 & 72.41 & \textbf{94.83} \\
          & & & Reference Diversity & 72.60 & 96.44 & \textbf{97.86} \\
          & & & Reference Satisfaction & 76.04 & 82.29 & \textbf{84.38} \\
        \cmidrule{2-7}
        & \multirow{4}{*}{R3} 
          & Richness & Claim Diversity & 77.69 & 87.40 & \textbf{96.07} \\
          \cmidrule{3-7}
          & & \multirow{3}{*}{\shortstack[l]{Concise/\\Usability}} & Answer Firstness & 80.47 & 95.05 & \textbf{97.66} \\
          & & & Answer Useful & 77.34 & 79.08 & \textbf{96.07} \\
          & & & Redundant Repetition & 78.01 & 85.86 & \textbf{98.43} \\
        \bottomrule
    \end{tabular}
    \vspace{-4pt}
\end{table*}

%% file: tables/table3.tex
\begin{table}[t]
    \centering
    \caption{Comparison of holistic preference alignment.}
    \vspace{-6pt}
    \label{tab:table3}
    \small
    \begin{tabular}{lc}
        \toprule
        Method & AUC \\
        \midrule
        GenRM & 70.90 \\
        Rubric & 72.13 \\
        Reward System (Ours) & \textbf{86.48} \\
        \bottomrule
    \end{tabular}
    \vspace{-10pt}
\end{table}

%% file: tables/table4.tex
\begin{table*}[t]
    \centering
    \caption{Comparison of different optimization methods across all reward dimensions.}
    \vspace{-4pt}
    \label{tab:table4}
    \small
    \begin{tabular}{lccccccc}
        \toprule
         & \multicolumn{2}{c}{Robustness (R1)} & \multicolumn{3}{c}{Bottom-line (R2)} & \multicolumn{2}{c}{Alignment (R3)} \\
        \cmidrule(lr){2-3} \cmidrule(lr){4-6} \cmidrule(lr){7-8}
        Method & Query & Evidence & Basic & Hallu & Format & Rich & Usability \\
        \midrule
        SFT & 0.9167 & 0.5809 & 0.9675 & 0.9176 & 0.9697 & 0.9222 & 0.7939 \\
        RFT & 0.9720 & 0.5880 & \textbf{0.9930} & 0.9260 & 0.9890 & 0.9600 & 0.8950 \\
        DPO & 0.9540 & 0.6120 & 0.9885 & 0.9510 & 0.9745 & 0.9690 & 0.8610 \\
        GRPO-GenRM & 0.9630 & 0.5980 & 0.9870 & 0.9340 & 0.9670 & \textbf{0.9840} & 0.8450 \\
        GRPO-Linear & 0.9636 & 0.5861 & 0.9906 & 0.9714 & 0.9730 & 0.9738 & 0.8604 \\
        GRPO-Gated & \textbf{0.9959} & \textbf{0.7089} & 0.9875 & \textbf{0.9836} & \textbf{0.9925} & 0.9832 & \textbf{0.9099} \\
        \bottomrule
    \end{tabular}
    \vspace{-8pt}
\end{table*}

%% file: figures/tex/fig4.tex
\begin{figure}[t!]
  \centering
  \includegraphics[width=\linewidth]{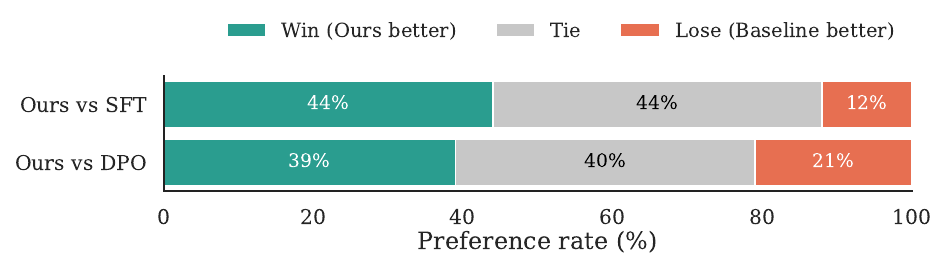}
  \vspace{-16pt}
  \caption{Comparison on generation quality of our policy against multiple baselines evaluated by human experts.}
  \vspace{-18pt}
  \label{fig:fig4}
\end{figure}

%% file: figures/tex/fig5.tex
\begin{figure*}[t!]
  \centering
  \includegraphics[width=\linewidth]{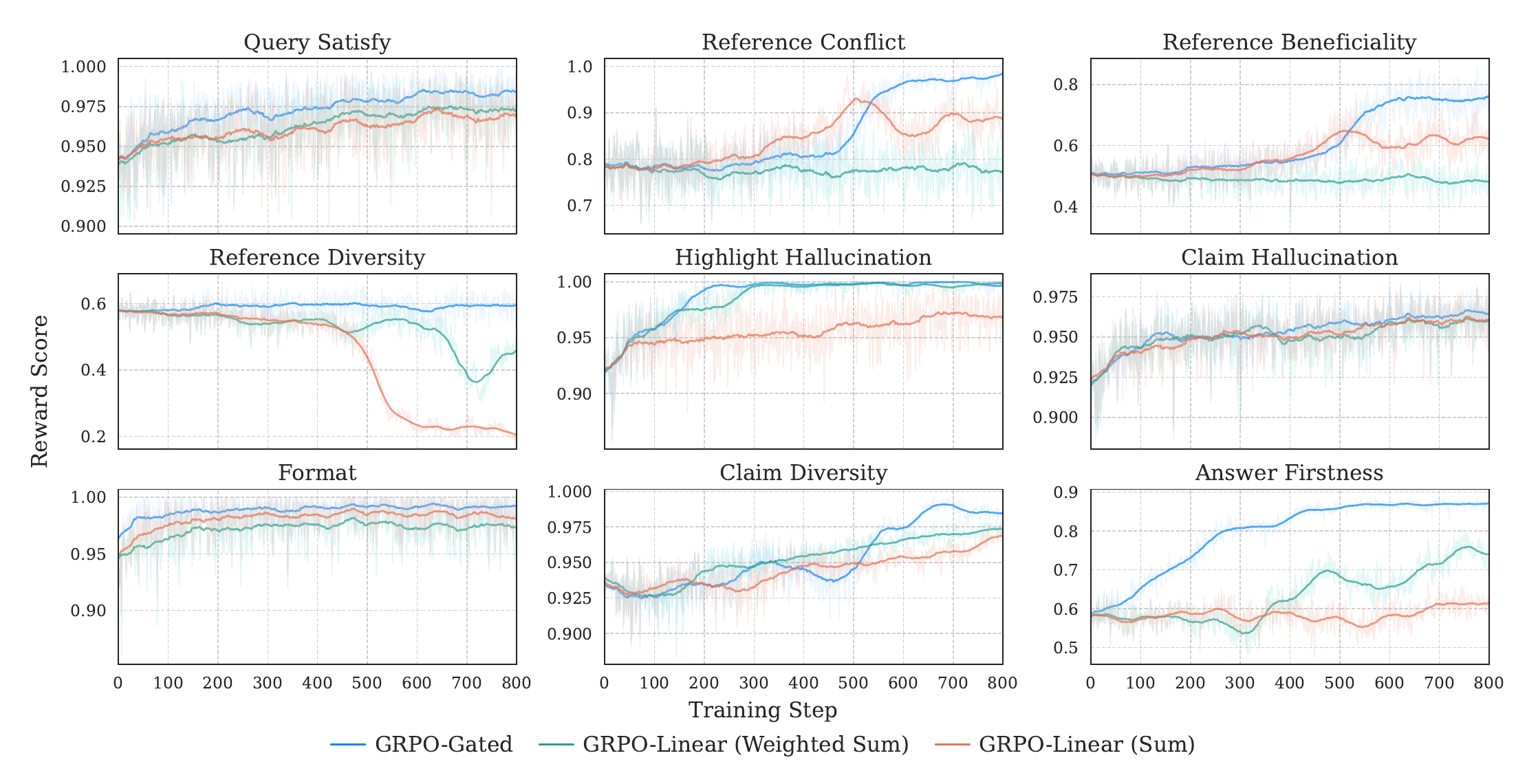}
  \vspace{-22pt}
  \caption{Training dynamics under different reward aggregation strategies. The curves illustrate the evolution of scores across distinct reward dimensions during training, comparing the Gated Aggregation strategy against the Linear baseline.}
  \vspace{-4pt}
  \label{fig:fig5}
\end{figure*}

%% file: figures/tex/fig6.tex
\begin{figure}[t!]
  \centering
  \includegraphics[width=\linewidth]{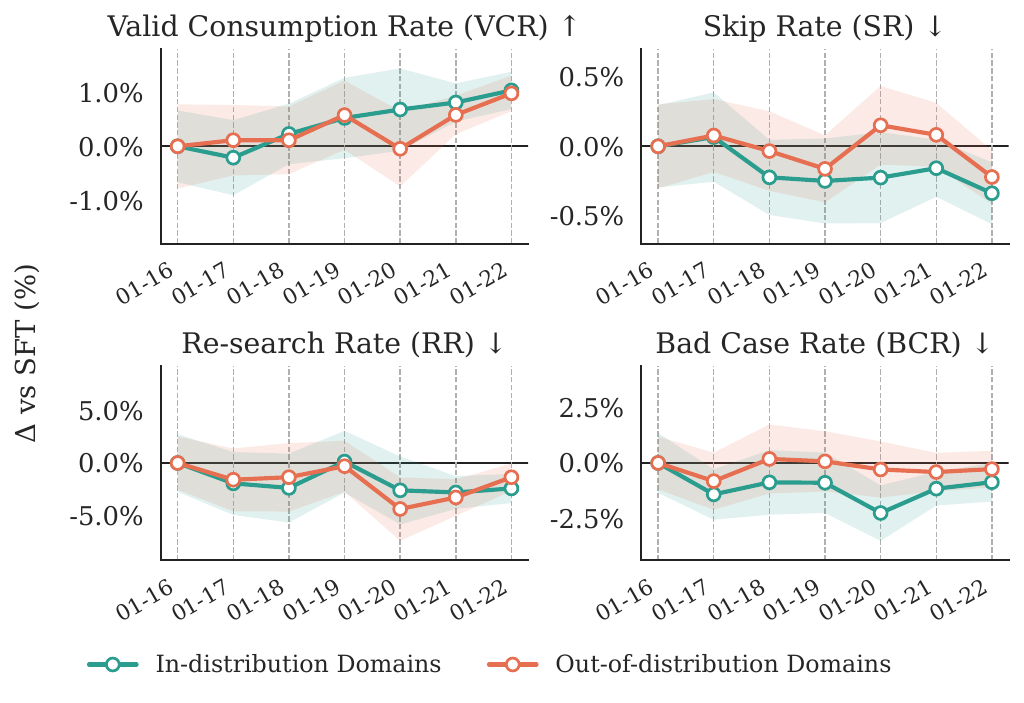}
  \vspace{-20pt}
  \caption{Results of the online A/B test on the RedNote platform conducted in 2026. The chart displays the relative changes in key user engagement metrics for our deployed model compared to the production baseline (SFT).}
  \vspace{-12pt}
  \label{fig:fig6}
\end{figure}

%% file: sections/5_conclusion.tex
\vspace{-4pt}
\section{Conclusion}
In this work, we introduced SearchLLM, an LLM optimized for open-ended generative search on large content platforms. We characterized the unique challenges of this setting, including robustness to noisy retrieval, safety guarantees, and alignment with diverse user needs. To solve these, we designed a multi-dimensional reward system that separates bottom-line constraints, such as factual grounding and format compliance, from behavioral objectives that drive user utility. Our approach uses the Gated Aggregation Strategy to decouple safety constraints from behavioral objectives, enabling effective policy optimization through Group Relative Policy Optimization (GRPO). The deployment of SearchLLM in RedNote, with over 150 million daily page views, was validated through online A/B tests. These tests showed improvements in generation quality and user engagement while maintaining safety standards. Our approach offers a scalable solution to align open-ended generative search systems with real-world constraints. Future work will extend this optimization to multi-modal contexts and incorporate personalized long-term memory to further refine the user experience.

%% file: sections/6_appendix.tex
\section{Qualitative Case Studies}
\label{sec:case_study}

To provide a concrete understanding of the performance improvements, we present three qualitative case studies from real-world traffic. For the sake of readability, all case studies are presented in English translation, while the original user queries and model responses were in Chinese. These examples compare the responses generated by SearchLLM against the SFT baseline:
\begin{itemize}
    \item \textbf{Robustness to Outdated Evidence (R1):} Figure~\ref{fig:case_temporal} (Case Study 1) demonstrates how our model effectively handles noisy retrieval by filtering out factually correct but temporally outdated information, whereas the baseline fails to recognize the event's expiration.
    \item \textbf{Bottom-line Safety Guarantees (R2):} Figure~\ref{fig:case_factuality} (Case Study 2) highlights our model's adherence to strict factual and safety constraints in medical domains, correcting the baseline's dangerous hallucination regarding pathogen classification.
    \item \textbf{Alignment with User Needs (R3):} Figure~\ref{fig:case_logic} (Case Study 3) illustrates how our model aligns with user intent for concise and logical solutions, eliminating the redundancy and logical conflicts observed in the baseline.
\end{itemize}

\section{Human Annotation Interface}
\label{app:annotation}
To ensure the high quality and consistency of our reward signals, we developed a specialized human annotation platform designed to align judges with expert preferences. As illustrated in Figure~\ref{fig:fig3}, the interface is structured to support a rigorous review process.

For each task, the annotator is presented with the raw user query and the candidate responses, alongside the intent analysis generated by the policy model and retrieved reference notes. This context ensures that judgments are grounded in accurate information rather than annotator intuition.

The annotation workflow consists of two granularities:
\begin{itemize}
    \item \textbf{Pointwise Evaluation:} Annotators first evaluate each response independently against our bottom-line constraints, scoring specific dimensions such as factual grounding, safety, and format compliance.
    \item \textbf{Pairwise Ranking:} After individual scoring, annotators perform a side-by-side comparison of candidate responses to provide holistic preference signals, which are closely correlated with the remaining reward dimensions.
\end{itemize}

This dual-granularity feedback mechanism enables hierarchical governance of non-negotiable safety boundaries and preference-oriented quality metrics during the reward modeling phase.

\section{Reward Dimension Definitions}
\label{app:reward}
In this section, we provide detailed definitions and implementation methods for the multi-dimensional reward system described in Section~\ref{sec:3_2}. Table~\ref{tab:table5} lists all criteria used in our hybrid evaluation stack, distinguishing between Bottom-line Constraints (Layer I) and Behavioral Objectives (Layer II). The "Implementation" column indicates whether a metric is computed via deterministic rules (Rule-based) or model judges (LLM-based). 

\section{Dataset Specifications}
\label{app:datasets}

To support training and evaluation, we construct four distinct datasets from RedNote search logs:

\begin{itemize}
    \item \textbf{Reward Training Dataset ($\mathcal{D}_{\text{RM-Train}}$):} Comprising 40,000 samples, this dataset facilitates the calibration of our hybrid evaluation stack and the training of baseline reward modeling methods. Constructed via stratified sampling of random logs and "hard samples" (historical failure cases), it adopts distinct formats: pairwise comparisons $\{(x, y_w, y_l)\}$ for subjective preference criteria, and pointwise instances $\{(x, y, s)\}$ with binary expert labels ($s \in \{0, 1\}$) for objective constraints.
    
    \item \textbf{Dimension-Specific Test Set ($\mathcal{D}_{\text{Eval-Dim}}$):} Designed to address RQ1, this held-out diagnostic set features samples explicitly tagged with specific quality issues (e.g., hallucination, formatting errors). Containing 800 to 3,600 examples per dimension, it allows us to evaluate the alignment with human judgment on individual criteria.
    
    \item \textbf{Holistic Preference Test Set ($\mathcal{D}_{\text{Eval-Holistic}}$):} Also serving RQ1, this dataset comprises 2,800 pairwise comparisons sourced from historical model checkpoints and human writing. It utilizes general "win/loss" labels to assess the holistic preference alignment of the reward system.
    
    \item \textbf{RL Optimization Dataset ($\mathcal{D}_{\text{RL}}$):} Comprising 500,000 unlabeled tuples $(q, h, E)$, this dataset supports policy optimization (RQ2--4). The training set mixes random traffic with user complaint queries to enhance robustness.
\end{itemize}

\section{Evaluation Metrics}
\label{app:metrics}

In this section, we provide the detailed definitions for the offline and online metrics used to assess our system.

\paragraph{\textbf{Offline Evaluation}}
We validate the reward model's alignment with human experts using two standard metrics:
\begin{itemize}
    \item \textbf{Accuracy (ACC):} Measures the agreement between reward signals and expert labels on binary constraint reward dimensions (e.g., Factuality, Format Compliance).
    \item \textbf{Area Under the Curve (AUC):} Evaluates the pairwise ranking capability on subjective preference dimensions (e.g., Richness, Usability). Given $N$ pairs $(y_w, y_l)$ where $y_w$ is the preferred response, and reward score $s(\cdot)$, the AUC is computed as:
    \begin{equation}
        \text{AUC} = \frac{1}{N} \sum_{i=1}^{N} \left[ \mathbb{I}(s(y_{w}^{(i)}) > s(y_{l}^{(i)})) + 0.5 \cdot \mathbb{I}(s(y_{w}^{(i)}) = s(y_{l}^{(i)})) \right].
    \end{equation}
\end{itemize}

\paragraph{\textbf{Online Evaluation}}
To measure real-world impact, we conducted online A/B testing and monitored the following metrics:
\begin{itemize}
    \item \textbf{Valid Consumption Rate (VCR):} The percentage of sessions where users dwell on the generated answer for a meaningful duration ($>5$s), indicating high utility.
    \begin{equation}
        \text{VCR} = \frac{1}{M} \sum_{j=1}^{M} \mathbb{I}(T_{\text{dwell}}^{(j)} > 5\text{s})
    \end{equation}
    \item \textbf{Skip Rate (SR):} The percentage of sessions where the answer is scrolled past immediately ($<1.5$s), serving as a proxy for irrelevance or poor formatting.
    \begin{equation}
        \text{SR} = \frac{1}{M} \sum_{j=1}^{M} \mathbb{I}(T_{\text{dwell}}^{(j)} < 1.5\text{s})
    \end{equation}
    \item \textbf{Re-search Rate (RR):} The frequency with which users issue a reformulated query immediately after viewing the result, indicating dissatisfaction.
    \item \textbf{Bad Case Rate (BCR):} The percentage of responses containing severe bottom-line violations (e.g., safety risks, obvious hallucinations). This metric is estimated via a daily human audit of sampled logs.
\end{itemize}

\section{Supplementary Training Dynamics}
We present the evolution curves for the remaining reward dimensions that were not visualized in the main text due to space constraints. As shown in Figure~\ref{fig:figb}, the Gated Aggregation Strategy demonstrates consistent stability across the full spectrum of reward signals, corroborating the conclusions drawn in Section~\ref{sec:4_4}.

\input{figures/tex/figa}
\input{figures/tex/fig3}
\input{tables/table5}
\input{figures/tex/figb}

%% file: figures/tex/figa.tex
\newcommand{\mycite}[1]{%
  \hspace{1pt}
  \textcolor{citeColor}{\scriptsize\texttt{[Doc~#1]}}%
}
\definecolor{citeColor}{HTML}{607D8B}

\begin{figure*}[t]
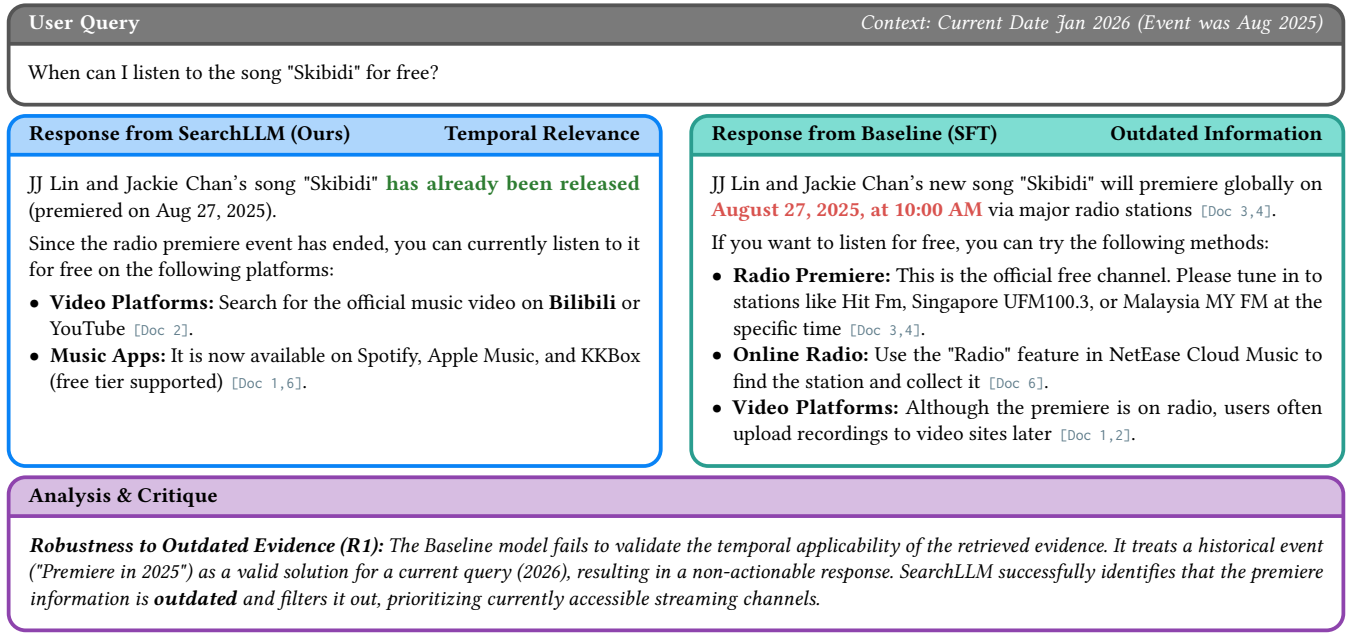

    \centering
    \small
    
    \begin{tcolorbox}[casestudy/query, title={\textbf{User Query} \hfill \normalfont\textit{Context: Current Date Jan 2026 (Event was Aug 2025)}}]
        When can I listen to the song "Skibidi" for free?
    \end{tcolorbox}
    
    \begin{minipage}[t]{0.49\textwidth}
        \begin{tcolorbox}[casestudy/ours, equal height group=group1, title={\textbf{Response from SearchLLM (Ours)} \hfill \textbf{Temporal Relevance}}]
            JJ Lin and Jackie Chan's song "Skibidi" \textcolor{textGreen}{\textbf{has already been released}} (premiered on Aug 27, 2025).
            
            Since the radio premiere event has ended, you can currently listen to it for free on the following platforms:
            \begin{itemize}[leftmargin=*, nosep]
                \item \textbf{Video Platforms:} Search for the official music video on \textbf{Bilibili} or YouTube\mycite{2}.
                \item \textbf{Music Apps:} It is now available on Spotify, Apple Music, and KKBox (free tier supported)\mycite{1,6}.
            \end{itemize}
        \end{tcolorbox}
    \end{minipage}
    \hfill
    \begin{minipage}[t]{0.49\textwidth}
        \begin{tcolorbox}[casestudy/baseline, equal height group=group1, title={\textbf{Response from Baseline (SFT)} \hfill \textbf{Outdated Information}}]
            JJ Lin and Jackie Chan's new song "Skibidi" will premiere globally on \textcolor{textRed}{\textbf{August 27, 2025, at 10:00 AM}} via major radio stations\mycite{3,4}.
            
            If you want to listen for free, you can try the following methods:
            \begin{itemize}[leftmargin=*, nosep]
                \item \textbf{Radio Premiere:} This is the official free channel. Please tune in to stations like Hit Fm, Singapore UFM100.3, or Malaysia MY FM at the specific time\mycite{3,4}.
                \item \textbf{Online Radio:} Use the "Radio" feature in NetEase Cloud Music to find the station and collect it\mycite{6}.
                \item \textbf{Video Platforms:} Although the premiere is on radio, users often upload recordings to video sites later\mycite{1,2}.
            \end{itemize}
        \end{tcolorbox}
    \end{minipage}
    
    \begin{tcolorbox}[casestudy/analysis]
        \textbf{Robustness to Outdated Evidence (R1):} The Baseline model fails to validate the temporal applicability of the retrieved evidence. It treats a historical event ("Premiere in 2025") as a valid solution for a current query (2026), resulting in a non-actionable response. SearchLLM successfully identifies that the premiere information is \textbf{outdated} and filters it out, prioritizing currently accessible streaming channels.
    \end{tcolorbox}
    
    \captionof{figure}{Case Study 1 highlighting improved temporal awareness for time sensitive queries}
    \label{fig:case_temporal}
\end{figure*}

\begin{figure*}[t]
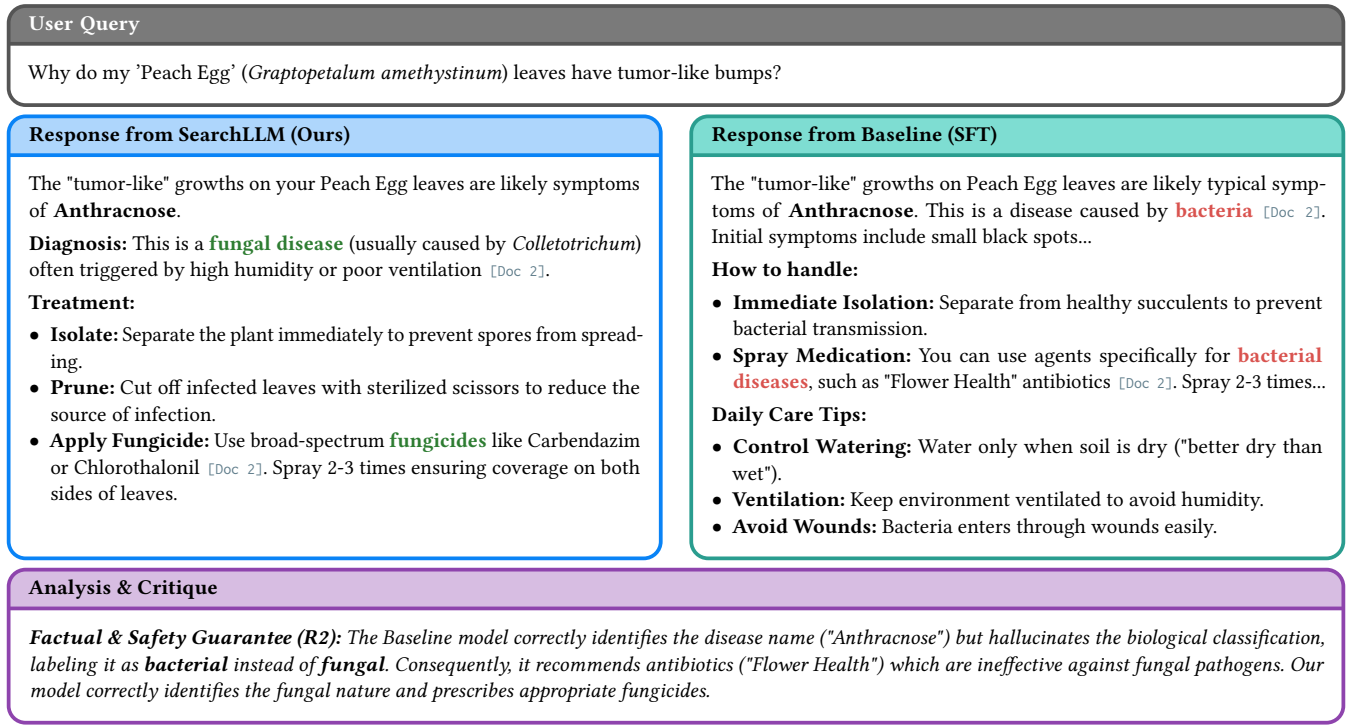

    \centering
    \small 
    
    \begin{tcolorbox}[casestudy/query, title={\textbf{User Query}}]
        Why do my 'Peach Egg' (\textit{Graptopetalum amethystinum}) leaves have tumor-like bumps?
    \end{tcolorbox}
    
    \begin{minipage}[t]{0.49\textwidth}
        \begin{tcolorbox}[casestudy/ours, equal height group=group2, title={\textbf{Response from SearchLLM (Ours)}}]
            The "tumor-like" growths on your Peach Egg leaves are likely symptoms of \textbf{Anthracnose}.
            
            \textbf{Diagnosis:} This is a \textcolor{textGreen}{\textbf{fungal disease}} (usually caused by \textit{Colletotrichum}) often triggered by high humidity or poor ventilation\mycite{2}.
            
            \textbf{Treatment:}
            \begin{itemize}[leftmargin=*, nosep]
                \item \textbf{Isolate:} Separate the plant immediately to prevent spores from spreading.
                \item \textbf{Prune:} Cut off infected leaves with sterilized scissors to reduce the source of infection.
                \item \textbf{Apply Fungicide:} Use broad-spectrum \textcolor{textGreen}{\textbf{fungicides}} like Carbendazim or Chlorothalonil\mycite{2}. Spray 2-3 times ensuring coverage on both sides of leaves.
            \end{itemize}
        \end{tcolorbox}
    \end{minipage}
    \hfill
    \begin{minipage}[t]{0.49\textwidth}
        \begin{tcolorbox}[casestudy/baseline, equal height group=group2, title={\textbf{Response from Baseline (SFT)}}]
            The "tumor-like" growths on Peach Egg leaves are likely typical symptoms of \textbf{Anthracnose}. This is a disease caused by \textcolor{textRed}{\textbf{bacteria}}\mycite{2}. Initial symptoms include small black spots...
            
            \textbf{How to handle:}
            \begin{itemize}[leftmargin=*, nosep]
                \item \textbf{Immediate Isolation:} Separate from healthy succulents to prevent bacterial transmission.
                \item \textbf{Spray Medication:} You can use agents specifically for \textcolor{textRed}{\textbf{bacterial diseases}}, such as "Flower Health" antibiotics\mycite{2}. Spray 2-3 times...
            \end{itemize}
            
            \textbf{Daily Care Tips:}
            \begin{itemize}[leftmargin=*, nosep]
                \item \textbf{Control Watering:} Water only when soil is dry ("better dry than wet").
                \item \textbf{Ventilation:} Keep environment ventilated to avoid humidity.
                \item \textbf{Avoid Wounds:} Bacteria enters through wounds easily.
            \end{itemize}
        \end{tcolorbox}
    \end{minipage}
    
    \begin{tcolorbox}[casestudy/analysis]
        \textbf{Factual \& Safety Guarantee (R2):} The Baseline model correctly identifies the disease name ("Anthracnose") but hallucinates the biological classification, labeling it as \textbf{bacterial} instead of \textbf{fungal}. Consequently, it recommends antibiotics ("Flower Health") which are ineffective against fungal pathogens. Our model correctly identifies the fungal nature and prescribes appropriate fungicides.
    \end{tcolorbox}
    
    \captionof{figure}{Case Study 2 demonstrating the correction of factual hallucinations in domain specific queries}
    \label{fig:case_factuality}
\end{figure*}

\begin{figure*}[t]
    \centering
    \small
    
    \begin{tcolorbox}[casestudy/query, title={\textbf{User Query}}]
        Which gate should I enter for Baixi from Wanda Resort?
    \end{tcolorbox}
    
    \begin{minipage}[t]{0.49\textwidth}
        \begin{tcolorbox}[casestudy/ours, equal height group=group3, title={\textbf{Response from SearchLLM (Ours)}}]
            From Wanda Resort to the Changbai Mountain \textbf{Baixi} scenic area, you should enter through the \textcolor{textGreen}{\textbf{Baixi Distribution Center Main Entrance}}.
            
            \textbf{Key Steps:}
            \begin{enumerate}[leftmargin=*, nosep]
                \item \textbf{Entrance:} Proceed directly to the main tourist hall for security check and ticket validation\mycite{1}.
                \item \textbf{Documents:} Must bring your ID card (Physical ID required) for pre-check\mycite{1}.
                \item \textbf{Timing:} Arrive 30 minutes before your booked slot\mycite{1}.
            \end{enumerate}
        \end{tcolorbox}
    \end{minipage}
    \hfill
    \begin{minipage}[t]{0.49\textwidth}
        \begin{tcolorbox}[casestudy/baseline, equal height group=group3, title={\textbf{Response from Baseline (SFT)}}]
            From Wanda to Changbai Mountain, it is usually recommended to enter from the \textcolor{textRed}{\textbf{West Slope Distribution Center}} main entrance\mycite{1}. This is the main pick-up point for buses.
            
            \textbf{Traffic and Cost:}
            \begin{itemize}[leftmargin=*, nosep]
                \item \textbf{Taxi:} From Wanda to Baixi is about 20 yuan more expensive than West Slope. Note: Taxis here do not use meters, keep the driver's phone number\mycite{1}.
            \end{itemize}
            
            \textbf{Tour Tips:}
            \begin{itemize}[leftmargin=*, nosep]
                \item \textbf{Motion Sickness:} The mountain road has 18 bends, prepare medicine\mycite{1}.
                \item \textbf{Clothing:} \textcolor{textRed}{\textbf{Even in summer}}, the wind is strong, so wear a windbreaker\mycite{1}.
            \end{itemize}
        \end{tcolorbox}
    \end{minipage}
    
    \begin{tcolorbox}[casestudy/analysis]
        \textbf{Alignment with User Needs (R3):} 
        1. \textbf{Entity Error:} The Baseline confuses "Baixi" with "West Slope Center". 
        2. \textbf{Redundancy:} It includes excessive irrelevant details (taxi scams, motion sickness) that dilute the answer. 
        3. \textbf{Logic Error:} It assumes the season is "Summer" without user input. Our model aligns with the user's specific need for navigation instructions.
    \end{tcolorbox}
    
    \captionof{figure}{Case Study 3 showcasing reduction of logical inconsistencies and redundancy in complex tasks}
    \label{fig:case_logic}
\end{figure*}

%% file: figures/tex/fig3.tex
\begin{figure*}[t!]
  \centering
  \includegraphics[width=\linewidth]{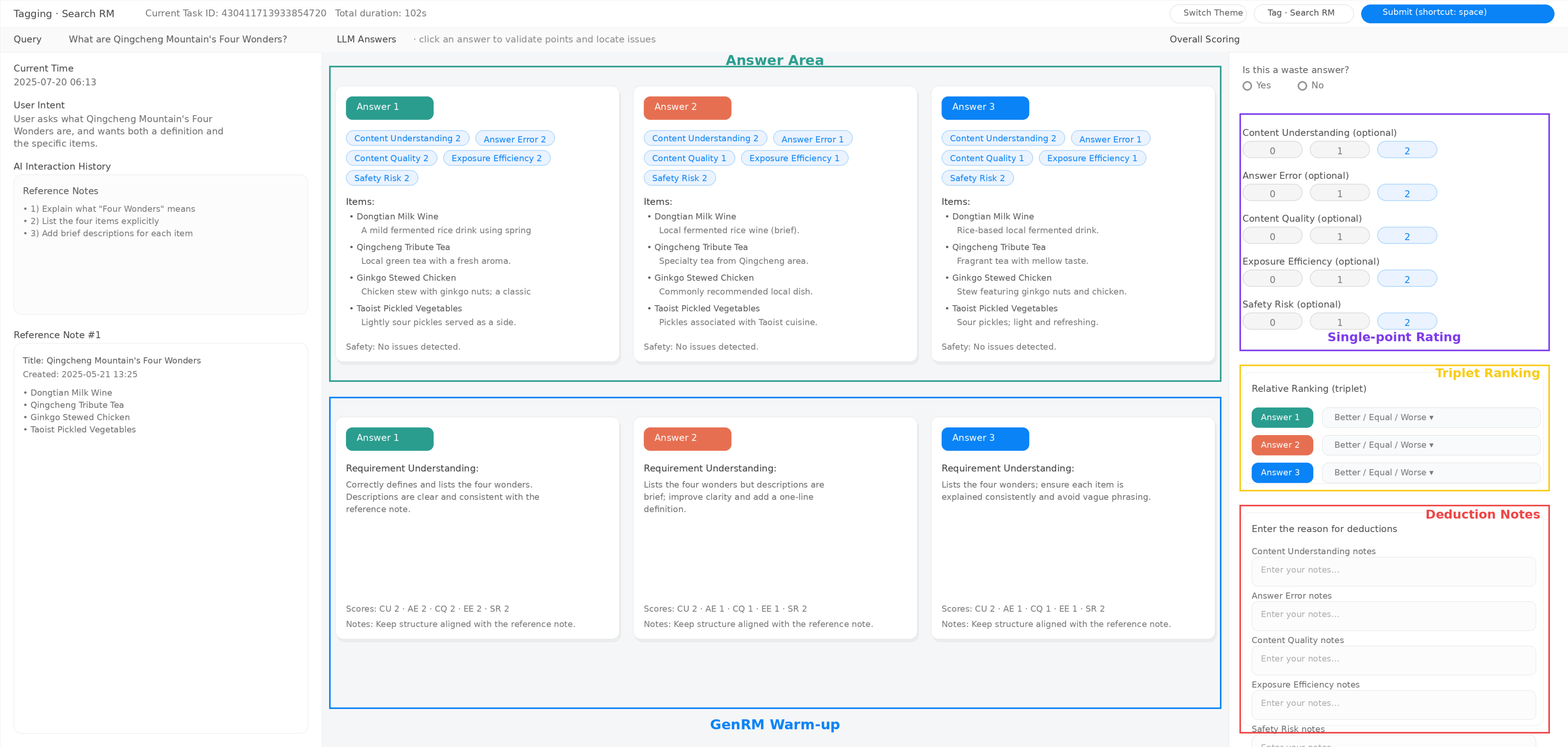}
  \caption{Screenshot of the human annotation interface. The workbench presents the query, user intent analysis, and reference notes. Annotators provide dual-granularity feedback: fine-grained scores on specific dimensions and holistic pairwise rankings.}
  \label{fig:fig3}
\end{figure*}

%% file: tables/table5.tex
\begin{table*}[t]
\centering
\caption{Detailed Definitions of Multi-Dimensional Reward Criteria.}
\label{tab:table5}
\resizebox{\textwidth}{!}{%
\begin{tabular}{l c l l l l}
\toprule
\textbf{Layer} & \textbf{Requirement} & \textbf{Subset} & \textbf{Reward Dimension} & \textbf{Implementation} & \textbf{Definition / Role} \\
\midrule
\multirow{9}{*}{\shortstack[l]{\textbf{Layer I}\\(Bottom-line)}} 
 & \multirow{9}{*}{R2} 
   & \multirow{4}{*}{Hallucination} 
     & Highlight Hallucination & LLM-based & First sentence free of obvious hallucinations. \\
 & & & Claim Hallucination & LLM-based & Non-hallucination at the claim level. \\
 & & & LLM Knowledge & LLM-based & Checks knowledge correctness with external validation. \\
 & & & No Supply Reject & LLM-based & Proper refusal behavior when no evidence is available. \\
 \cmidrule(l){3-6}
 & & \multirow{3}{*}{Basic Quality} 
     & Self Consistency & LLM-based & Logical self-consistency, no internal contradictions. \\
 & & & Answer Quality & LLM-based & Avoids gibberish or extremely low-quality answers. \\
 & & & Multi-round Repeat Conflict & LLM-based & No strong repetition or contradiction across turns. \\
 \cmidrule(l){3-6}
 & & \multirow{2}{*}{Format} 
     & Format & Rule-based & Enforces structured and well-formed output format. \\
 & & & Response Length & Rule-based & Keeps answer length in an acceptable range. \\
\midrule
\multirow{11}{*}{\shortstack[l]{\textbf{Layer II}\\(Behavioral)}} 
 & \multirow{7}{*}{R1} 
   & \multirow{2}{*}{Query} 
     & Query Satisfaction & LLM-based & Basic intent alignment, avoids “answering B when asked A”. \\
 & & & Planning Quality & LLM-based & Quality of reasoning and planning under uncertainty. \\
 \cmidrule(l){3-6}
 & & \multirow{5}{*}{Evidence} 
     & Reference Beneficiality & LLM-based & Evidence truly helps answer the question. \\
 & & & Reference Conflict & LLM-based & Detects strong conflicts among evidence items. \\
 & & & Reference Irrelevant & LLM-based & Detects evidence completely irrelevant to the query. \\
 & & & Reference Satisfaction & LLM-based & Evidence covers primary and secondary needs. \\
 & & & Reference Diversity & LLM-based & Promotes diverse evidence for robustness. \\
 \cmidrule(l){2-6}
 & \multirow{4}{*}{R3} 
   & Richness 
     & Claim Diversity & LLM-based & Rich and diverse answer claims. \\
 \cmidrule(l){3-6}
 & & \multirow{3}{*}{Usability} 
     & Answer Useful & LLM-based & Low fraction of off-topic content; conciseness. \\
 & & & Answer Firstness & LLM-based & Places core answer early for quick consumption. \\
 & & & Redundant Repetition & LLM-based & Controls severe semantic redundancy and rambling. \\
\bottomrule
\end{tabular}%
}
\end{table*}

%% file: figures/tex/figb.tex
\begin{figure*}[t!]
  \centering
  \includegraphics[width=\linewidth]{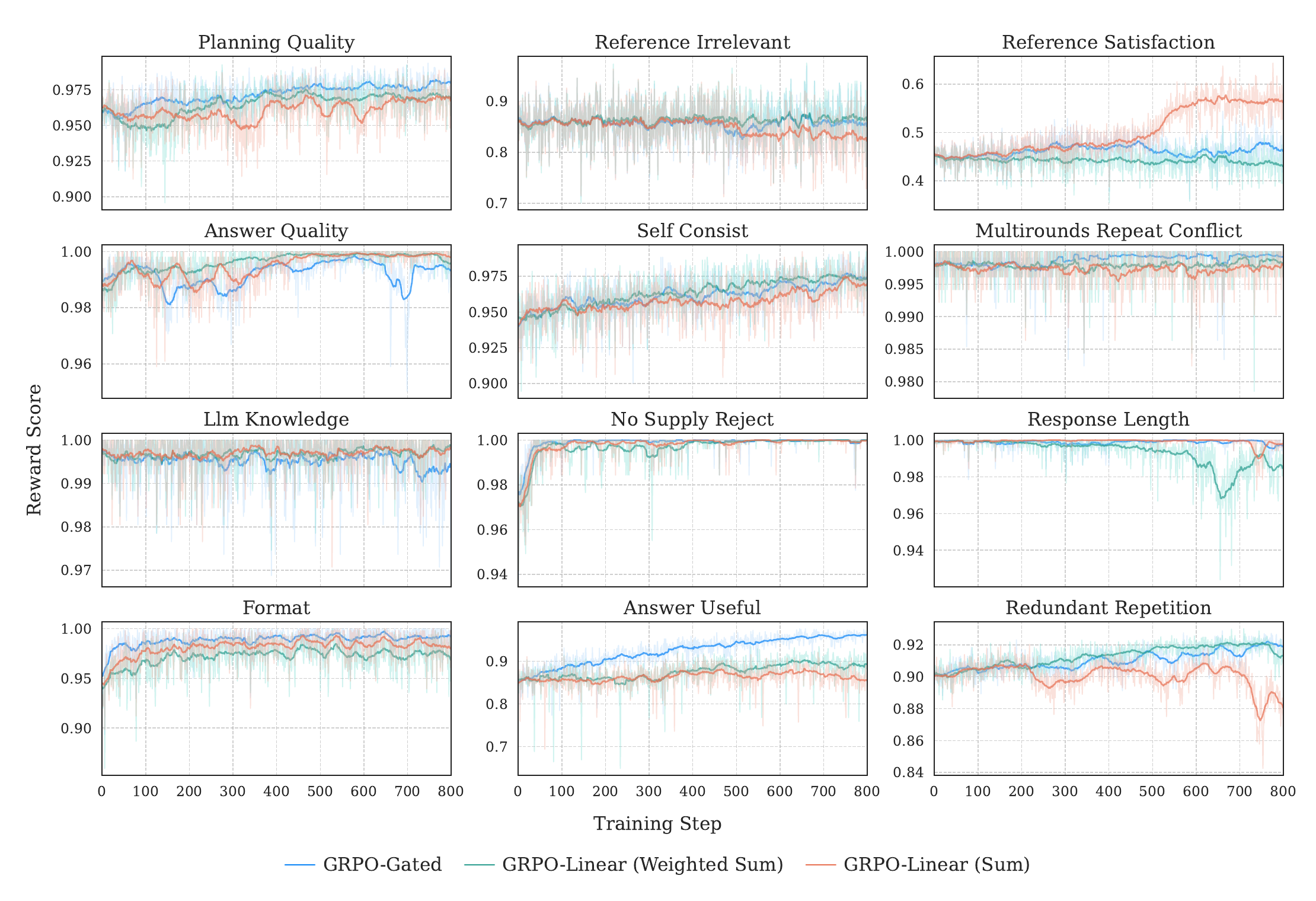}
  \caption{Supplementary training dynamics showing the evolution of the remaining reward dimensions.}
  \label{fig:figb}
\end{figure*}